%% file: amlc_2022.tex
\documentclass{article}

\usepackage[preprint,nonatbib]{amlc_2022}

\usepackage[utf8]{inputenc} 
\usepackage[T1]{fontenc}    
\usepackage{hyperref}       
\usepackage{url}            
\usepackage{booktabs}       
\usepackage{amsfonts}       
\usepackage{nicefrac}       
\usepackage{microtype}      
\usepackage{xcolor}         
\usepackage[colorinlistoftodos]{todonotes} 

\newcommand{\vi}[1]{\textbf{\color{red} VI: #1}}

\usepackage{color}
\usepackage{array}
\usepackage{amsfonts}
\usepackage{xspace}
\usepackage{amsmath}
\usepackage{tikz}
\definecolor{q1}{RGB}{208,2,27}
\definecolor{q2}{RGB}{126,211,33}
\definecolor{p1}{RGB}{208,2,27}
\definecolor{p2}{RGB}{208,2,27}
\definecolor{p3}{RGB}{126,211,33}

\usepackage{graphicx} 
\usepackage{pgfplots}
\usepackage{subcaption}
\usepackage{ntheorem}
\graphicspath{ {./figures/} }
\usepackage{multirow}
\usepackage{comment}
\usepackage{bm}
\usepackage{etoolbox}
\preto\tabular{\setcounter{magicrownumbers}{0}}
\newcounter{magicrownumbers}
\newcommand\rownumber{\stepcounter{magicrownumbers}\arabic{magicrownumbers}}

\usepackage[flushleft]{threeparttable}

\newcommand{\mathbbm}[1]{\text{\usefont{U}{bbm}{m}{n}#1}}
\newcommand{\sys}{LM-GNN\xspace}
\newcolumntype{H}{>{\setbox0=\hbox\bgroup}c<{\egroup}@{}}

\author{%
  David S.~Hippocampus\thanks{Use footnote for providing further information
    about author (webpage, alternative address)---\emph{not} for acknowledging
    funding agencies.} \\
  Department of Computer Science\\
  Cranberry-Lemon University\\
  Pittsburgh, PA 15213 \\
  \texttt{hippo@cs.cranberry-lemon.edu} \\
}
\title{Efficient and effective training of language and graph neural network models}
\author{Vassilis~N.~Ioannidis, Xiang~Song, Da~Zheng, George~Karypis
\\Amazon Web Services AI, USA \AND  Houyu~Zhang, Jun~Ma, Yi~Xu, Belinda~Zeng, Trishul~Chilimbi
\\Amazon Search AI, USA}

\begin{document}

\maketitle
\begin{abstract}
Can we combine heterogenous graph structure with text to learn high-quality semantic and behavioural representations? Graph neural networks (GNN)s encode numerical node attributes and graph structure to achieve impressive performance in a variety of supervised learning tasks. Current GNN approaches  are challenged by textual features, which typically need to be encoded to a numerical vector before provided to the GNN that may incur some information loss. 
In this paper, we put forth an efficient and effective framework termed  language model GNN (\sys) to jointly train large-scale language models and graph neural networks. 
The effectiveness in our framework is achieved by applying stage-wise fine-tuning of the BERT model first with heterogenous graph information and then with a GNN model. 
Several system and design optimizations are proposed to enable scalable and efficient training.
\sys accommodates node and edge classification as well as link prediction tasks.  We evaluate the \sys framework in different datasets performance and showcase the effectiveness of the proposed approach. \sys provides competitive results in an Amazon query-purchase-product application.
\end{abstract}
\section{Introduction}
GNNs rely on a layered processing architecture comprising trainable graph convolutional operations to linearly combine features per graph neighborhood, followed by pointwise nonlinear functions applied to the linearly transformed features~\cite{bronstein2017geometric}. GNNs have shown remarkable success in a variety of graph machine learning tasks both in supervised and unsupervised learning settings~\cite{hamilton2017representation}. 
Typically, the graphs used for profiling GNN models have node features as numerical attributes. These numerical attributes may be the output of network that encodes a much richer original information that is in the form of text or picture. One could apply such a general pre-trained network to extract the represenations and use them as feature vectors in a GNN. However, as we detail in this work such an approach is not optimal. This raises the question of how can we train better GNN models with rich text features.
This work presents a stage-wise fine-tuning framework termed \sys for encoding text data with transformers and GNN models.

We implement a sequence of fine-tuning steps to gradually infuse the transformer model with graph structure information. Our study reveals the necessity of pre-fine-tuning the transformer for graph-aware tasks. The graph-aware BERT is subsequently fine-tuned together with a GNN model for performing any downstream tasks, which allows the model to access multi-hop information. Our stage-wise fine-tuning, besides achieving good performance, significantly reduces training time compared to end-to-end training without our stage-wise approach because the mode converges faster. Further, \sys is a distributed framework which can scale to hundreds of millions nodes. Besides improving the model performance we implement a series of system and design optimizations to speed up the overall training speed. Our \sys framework improves compared to baseline performance in four public datasets and one query-purchase-product application.

\section{Related work}
Node classification is typically formulated as a semi-supervised learning (SSL) task over graphs, where the labels for a subset of nodes are available for training~\cite{belkin2004regularization}. GNNs achieve state-of-the-art performance in SSL by utilizing regular graph convolution~\cite{kipf2016semi} or graph attention~\cite{velivckovic2017graph}, while these models have later been extended in the heterogeneous graph setting~\cite{schlichtkrull2018modeling,fu2020magnn,wang2019heterogeneous}. Similarly, another prominent graph machine learning task is link prediction with numerous applications in recommendation systems~\cite{wang2017knowledge} and drug discovery~\cite{zhou2020network,ioannidis2020few}.  Knowledge-graph (KG) embedding models for link prediction rely on mapping the nodes and edges of the KG to a vector space by maximizing a score function for existing KG edges~\cite{wang2017knowledge,yang2014embedding,zheng2020dgl}. Relational graph convolutional network (RGCN) models~\cite{schlichtkrull2018modeling} have been successful in link prediction and contrary to KG embedding models can further utilize node features. The aforementioned graph machine learning tasks may also be addressed using unsupervised learning approaches. Graph representation learning approaches map nodes in an embedding space where the graph topological information and structure is preserved~\cite{hamilton2017representation}. Such unsupervised representations can be consumed by any downstream model for task-specific prediction.  Unsupervised methods employ a GNN encoder that generates node embedding and is supervised by a task dependent decoder. Typically decoders perform matrix factorization~\cite{tang2015line,ahmed2013distributed,belkin2002laplacian,cao2015grarep,ou2016asymmetric}, random walks~\cite{grover2016node2vec,perozzi2014deepwalk}, or a combination of learning tasks~\cite{ioannidis2020panrep}. Language models (LM)s are powerful in modeling text data~\cite{devlin2018bert}. Harnessing the power of LMs with graph data is under-explored. This work details a framework for training large-scale LMs jointly with GNNs. 
Recent work~\cite{chien2021node} also identifies that pre-training BERT models in graph data can be beneficial and exploits a neighborhood prediction objective to enrich the BERT model with graph information. However, this work~\cite{chien2021node} did not explore to fine-tune the BERT and GNN model together. Another prominent work in~\cite{li2021adsgnn,zhu2021textgnn} trains GNN models for improving the search results in sponsored search. The work there can be seen as a special case of this framework, although~\cite{li2021adsgnn,zhu2021textgnn} did not explore the stage-wise fine-tuning that we introduce in this work.

\section{Definitions and Problem formulation}
\label{sec:prob}


A heterogeneous graph with $T$ node types and $R$ relation types is defined as $\mathcal{G}:=\{\{\mathcal{V}_{t}\}_{t=1}^{T},\{\mathcal{E}_r\}_{r=1}^{R}\}$. 
The node types represent the different entities and the relation types represent how these entities are semantically associated to each other. For example, in the query-product network of Fig.~\ref{fig:querygraph}, the node types correspond to queries and products, whereas the relation types may correspond to \emph{whether a product was clicked based on a query} and \emph{whether a product was purchased based on a query} relations. The number of nodes of type $t$ is denoted by $N_t$ and its associated node set by  $\mathcal{V}_{t}:=\{n_t\}_{n=1}^{N_t}$. The total number of nodes in $\mathcal{G}$ is $N:=\sum_{t=1}^T{N_t}$ and the total number of edges is $E:=\sum_{r=1}^R|\mathcal{E}_{r}|$ where $||$ denotes the number of elements in the set. The $r$th relation type, $\mathcal{E}_{r}:=\{(n_{t}, r ,n'_{t'})\in\mathcal{V}_{t}\times \mathcal{V}_{t'} \}$,  holds all interactions of a certain type among $\mathcal{V}_{t}$ and $\mathcal{V}_{t'}$ and may represent that a product is \emph{was-clicked-based} on a query.

Each node $n_t$ is also associated with a short text. In the query-product network for example, each product is accompanied by a title and each query by the query text. Notice that different node types could have texts that are drawn from different distribution, e.g., the text for titles can be distinctly different from that for queries. Oftentimes, such text features are mapped to a $F\times1$ embedding vector $\mathbf{x}_{n_t}$ by transformers in a task independent fashion. In this paper, we will explore different methods to implement this mapping, that directly relate to the downstream application of interest.

\section{\sys Models : Adapt and fine-tune}
In this paper, our high-level goal is to investigate how to fuse transformer and GNN models to learn informative representations from graph and text data.  Our \sys framework achieves this by stage-wise fine-tuning that gradually fuses the transformer with graph information. 

\subsection{Semantic  encoder} 
We employ the BERT model~\cite{devlin2018bert} as the transformer in the \sys framework to encode the nodes textual semantics. Given a node's text BERT encodes the textual information to a $F\times1$ embedding vector $\mathbf{x}_{n_t}$. This embedding vector corresponds to the [CLS] token embedding of the BERT model and the mapping from the text to the embedding is defined as $\mathbf{x}_{n_t} := g_{\text{\tiny{BERT}}}(n_t; \mathbf{W}_\text{\tiny{BERT}})$. The mapping is controlled by the learnable parameters $\mathbf{W}_\text{\tiny{BERT}}$. 
These parameters can be instantiated by any language model pre-training approach, e.g., masked language modeling (MLM).
Pre-training BERT models on large unlabeled text data has shown significant benefit in different LM applications. However, transferring this benefit for graph ML applications is not straightforward. We employ the techniques in Sec.~\ref{sec:strucpred} to pre-train BERT models with graph data. For different node-types $t\in\{1,\ldots,T\}$ in the graph  we may consider different semantic encoders; e.g. queries and products.

\subsection{Graph Encoder}
\label{sec:encoder}

Although the \sys framework can utilize any GNN model as an encoder~\cite{wu2020comprehensive}, in this paper \sys uses a modified RGCN encoder~\cite{schlichtkrull2018modeling}. RGCNs extend the graph convolution operation~\cite{kipf2016semi} to heterogeneous graphs. The $l$th self-RGCN layer computes the $n$th node representation $\mathbf{h}^{(l+1)}_{n}$ as follows
\begin{align}
\mathbf{h}^{(l+1)}_{n}
	:=
	\sigma\left(\mathbf{W}_{\text{self}}^{(l)}{\mathbf{h}}_{n}^{(l)}+\sum_{r=1}^{R}\sum_{ n'\in\mathcal{N}_n^{r}} 
	 \mathbf{W}^{(l)}_{r}{\mathbf{h}}_{n'}^{(l)} \right),
	\label{eq:sem}
\end{align}
where $\mathcal{N}_n^{r}$ is the neighborhood of node $n $ under relation $r$, $\sigma$ the rectified linear unit non linear function,  $\mathbf{W}^{(l)}_{r}$ is a learnable matrix associated with the $r$th relation, and $\mathbf{W}_{\text{self}}^{(l)}$ is a projection matrix for the nodes embedding in layer $l$. 
Our adaptation augments the messages at each layer with a projected transformation of the node embedding for that layer acting as a self loop with an appropriate weight matrix. This adaptation over the traditional RGCN the node's own representation based on the semantic encoder is  essential for the downstream applications and needs to be treated separately by the model.
The matrix $\mathbf{H}$ in this paper represents the embedding extracted in the final layer. The node features are the input of the first layer in the model i.e., $\mathbf{h}^{(0)}_{n}=\mathbf{x}_n$, where $\mathbf{x}_n$ is the node feature for node $n$. 


\subsection{Supervision approaches} 

\textbf{Structure prediction task}.\label{sec:strucpred}
Link prediction is a specific type of structure prediction, which will be our focus in the following section. Consider the heterogeneous graph  $\mathcal{G}$ in Sec.~\ref{sec:prob}.  Given the sets of links $\{\mathcal{E}_{r}\}_{r=1}^{R}$, and the node features the goal of link prediction is to predict whether a new set of node pairs are linked or not.

Typically, structure prediction models utilize a contrastive loss function that requires the model to distinguish among positive and negative examples~\cite{zheng2020dgl}. In this context, positive examples are the set of existing links in the graph. The negative examples, which are links that the model should classify as nonexistent, are typically sampled from the missing links in the graph. For each \emph{positive triplet} $q=(n_t,{r},{n'}_{t'})$ a number of negative links is generated by corrupting the head and tail entities at random $(n_t,{r},{\nu'}_{t'})$ and $(\nu_t,{r},{n'}_{t'})$. 

The minimization function for link prediction can be defined as follows
\begin{align}
\label{eq:rgcnsup}
\sum_{(n_t,{r},{n'}_{t'}) \in \mathbb{D}^+ \cup \mathbb{D}^-} \log(1+\exp(-y \times
c(n_t,r,{n'}_{t'})),
\end{align}
where $c$ is a scoring function that return as scalar given the head, and tail nodes and the relation such as the DistMult model~\cite{yang2014embedding}, $\mathbb{D}^+$ and $\mathbb{D}^-$ are the positive and negative sets of triplets and $y=1$ if the triplet corresponds to a positive example and $-1$ otherwise.


\textbf{Node prediction task}.
Each node ${n}$ has a label $y_{n}\in\{0,\ldots,P-1\}$, which in the query-product network may represent the type of a product. In semi-supervised learning, we know labels only for a subset of nodes $\{y_{{n}}\}_{{n}\in\mathcal{M}}$, with $\mathcal{M} \subset\mathcal{V}$. This partial availability may be attributed to privacy concerns (medical data); energy considerations  (sensor networks); or unrated items (recommender systems). The ${N}\times P$ matrix  $\mathbf{Y}$ is the one-hot representation of the true node labels; that is, if $y_{n}=p$ then $Y_{{n}p}=1$ and $Y_{{n}p'}=0, \forall p'\ne p$. 
The minimization objective in this task is the cross-entropy loss.
%
%

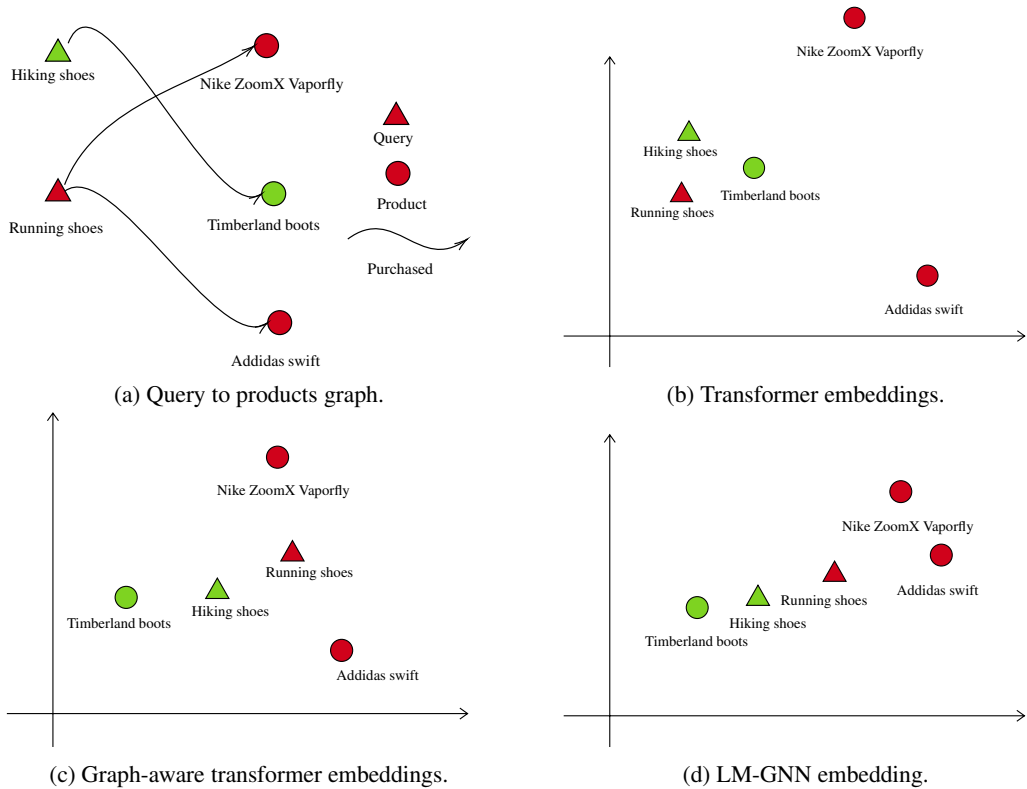
\begin{figure*}
  \centering
  \begin{subfigure}[t]{0.47\columnwidth}
    \resizebox{.95\linewidth}{!}{\input{figures/embedding_analysis/basic_graph}}
    \caption{Query to products graph.}\label{fig:querygraph}
    \end{subfigure}
    \hfill
    \begin{subfigure}[t]{0.47\columnwidth}
        \resizebox{.95\linewidth}{!}{\input{figures/embedding_analysis/tembed}}
        \caption{Transformer embeddings.}\label{fig:tremb}
    \end{subfigure}
    \hfill
    \begin{subfigure}[t]{0.47\columnwidth}
        \resizebox{.95\linewidth}{!}{\input{figures/embedding_analysis/gtembed}}
        \caption{Graph-aware transformer embeddings.}\label{fig:gtremb}
    \end{subfigure}
    \hfill~
    \begin{subfigure}[t]{0.47\columnwidth}
        \resizebox{.95\linewidth}{!}{\input{figures/embedding_analysis/gnnembed}}
        \caption{\sys embedding.}\label{fig:glm}
    \end{subfigure}
    \caption{(a) The underlying graph among products and queries where an edge signifies that a query leads to the purchase of a product.  (b-d) The 2-D projected embeddings as generated by different pipelines.  (b) The transformer maps entities solely on text and fails to capture semantic similarity, besides language based e.g., shoes are close to boots. (c) The graph-aware transformer maps connected entities close, however fails at capturing higher order relations and embeds the two running shoes in different regions. (d)The proposed \sys captures the connectivity, higher order structure, as well as text semantics and provides a refined representations useful for retrieval tasks.}
    \label{fig:embed_plots}
\end{figure*}

\textbf{Edge prediction task}.
Each link $l$ of a certain type ${r}$ may also associated with a label of interest $\psi_{l}\in\{0,\ldots,\Pi-1\}$. For example, consider the query to product graph in Fig.~\ref{fig:querygraph}, where the task becomes to predict whether a pair of connected (query, product) is an exact search match or not. Hence, here given an existing link we predict the class label, which is different from the structure prediction task in Sec.~\ref{sec:strucpred} where the objective is to predict the existence of a link. 

The final predicted label for link $l$ is the output of the following edge classification decoder
\begin{align}
\hat{\psi_{l}}=\mathbf{W}_{\text{ec}}(\mathbf{h}_{n_t}\mathbin\Vert\mathbf{h}_{{n'}_{t'}})
    \label{eq:eclabel}
\end{align}
where $\mathbf{W}_{\text{ec}}$ is a projection matrix of appropriate dimension and $\mathbf{W}_{\text{ec}}$ denotes concatenation. Hence, the predicted label for $l$ is a function of the entity embeddings for the nodes at the endpoints of the link ${n_t}, {n'}_{t'}$. 

\subsection{\sys: Training at scale using graph and text}

A straightforward approach would directly use the LM encoder as a semantic encoder that feeds representations to the GNN encoder, and train such an architecture in an end-to-end fashion. However, training large scale language models and graph neural networks involves challenges relating to efficiency and effectiveness. 

Effectiveness challenges stem from the fact that the pre-trained language model is well optimized in language tasks but has not trained before in graph tasks, which surfaces three main issues. (1) Using such pre-trained transfomers may not be the most appropriate initialization and may trap the GNN to a sub-optimal local minimum. (2) {Further, the well optimized transformer for the text tasks, may be more resistant in parameter updates.} (3) Another hurdle stems from the random initialization of the GNN weights relative to the well-attuned transformer model, which may challenge the optimization of such an end-to-end framework.

Efficiency challenges relate to the large number of neighbors required by message passing in GNNs. In mini-batch training of a $k$-layer GNN the $k$-hop ego-network of every target node is created and the target node embedding is  computed as a function of all the node in the expanded ego-network (also known as source nodes). 
The number of source nodes  in an ego-network may be very large even for shallow GNNs. Alleviating this issue, recent GNN approaches apply random sampling~\cite{graphsage} to reduce the number of neighbors. However, even with a shallow GNN (2 layers) and modest sampling (20 neighbors per layer), there are up to 400 source nodes. This remains a serious challenge in our unique setting where the transformer model needs to make 400 forward passes to calculate the embedding of a single target node. As a consequence the size of the required GPU is quite large even for small mini-batch sizes, which is a unique challenge in our framework.

\subsubsection{Addressing effectiveness}
Consider the search graph among products and queries depicted in Fig.~\ref{fig:querygraph}. Such a graph is typically encountered in catalog systems for query-product datasets. 
One could attempt to directly use the embedding generated by a transformer as an input to a GNN model for further fine-tuning. However, the transfomer embedding of such a model will only take into account the text information and may introduce noise at message passing. Indeed, Fig.~\ref{fig:tremb} shows that embeddings that are connected in the graph may be located in different regions of the embedding space. The poor performance of such a scheme is also detailed in the experiments; see Section~\ref{sec:exp}.
Our contribution in this context is to pre-fine-tune the transformer with graph information, which will endow the text embeddings with relational semantics and boost the performance when used as a semantic encoder. 

\begin{figure*}
  \centering
  \resizebox{1.05\linewidth}{!}{\input{figures/models/gtrans}}
  \hrulefill\vspace{15pt} 
  \medskip
\hrulefill\par
    \resizebox{1.05\linewidth}{!}{\input{figures/models/lmgnn}}
    \caption{(Top) The graph-aware transformer framework relies on the input text to predict whether two entities are connected in the heterogenous graph.  (Bottom) The \sys framework employs the graph-aware transformer as a semantic encoder that is further fine-tuned using the GNN encoder, for predicting links in the heterogeneous graph. Different than the graph-aware transformer framework the \sys can access nodes in multi-hop neighborhood.}
    \label{fig:framework}
\end{figure*}
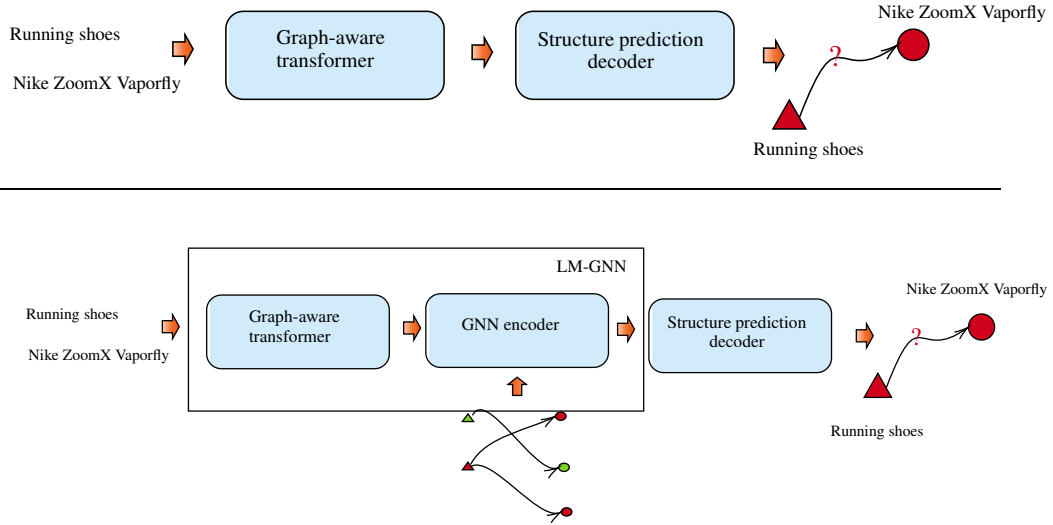

\textbf{Graph-aware pre-fine-tuning.}  We consider the structure prediction decoder that  directly uses the scoring function $c$ instantiated in Section~\ref{sec:strucpred}. The graph-aware transformer model directly uses the structure prediction decoder as a supervision to predict whether an edge exists among two nodes or not. Specifically, the transformer generates the CLS token embeddings for the text associated with the nodes and the vectors are contributing to the loss~\eqref{eq:rgcnsup}. The resulting graph-aware transformer embeddings respect both the semantics introduced by the language as well as the relations imposed by the graph; see also Fig.~\ref{fig:gtremb}.
The graph-aware pre-fine-tuning also results LM that is more suitable for the end-to-end training with the GNN, which is also supported in Section~\ref{sec:exp}. The top part of Fig.~\ref{fig:framework} showcases the graph-aware pre-fine-tuning pipeline.

Our proposed framework~\sys employs the graph-aware transformer as a semantic encoder that first embeds the text and then is fed to the GNN encoder.  However, since the GNN model is typically initialized at random this may challenge the end-to-end fine-tuning method and get trapped in not desirable local minima.  Hence, we warm-start the GNN weights by keeping fixed the transformer weights for a few iterations and optimize only the GNN encoder. This way we can have a good initialization for the GNN model weights before we attempt the joint training. Finally, we fine-tune end-to-end the semantic encoders and GNN models for the downstream task. Such a scheme will provide a good initial point for the GNN model. The resulting embeddings abide by the text semantics, graph relations and the multi-hop graph structure; see Fig.~\ref{fig:glm}.   Our overall stage-wise fine-tuning pipeline is depicted in the bottom part of Fig.~\ref{fig:framework}. 

\subsubsection{Addressing efficiency}
The high computation overhead and memory consumption required by the \sys framework limits the wide applicability of the approach. Addressing these issues, we adopt several optimizations to efficiently train \sys.

\textbf{Back-propagate on samples}.
Instead of back-propagating gradients to the transformer models on all nodes, we sub-sample a fixed-size number of nodes (train nodes) in a mini-batch where we back-propagate gradients to the BERT model. For the rest of the nodes (inference nodes), we just run BERT forward computation to generate BERT embeddings. To further reduce memory consumption and allow training in limited GPU machines, we split the inference nodes into multiple sub-batches of fixed size.

\textbf{Cache BERT embeddings}. To further reduce transformer computations, we cache text embeddings of some nodes in a mini-batch. During the training, whenever we compute new text embeddings, we save them in the cache. Whenever we need node BERT embeddings, we fetch them from the cache. 
Some cached text embeddings may be out-of-date in a large graph, which may lower the overall model accuracy. 


\textbf{Joint negative sampling}. Link prediction task requires positive and negative samples to be trained as detailed in Sec.~\ref{sec:strucpred}. Hence, we construct $k$ negative edges for each positive edge in a mini-batch. By default, we sample $k$ negative edges for each positive edge independently, which requires us to sample $k \times n$ new nodes with $n$ is the number of positive links. The optimization for joint negative sample samples $n$ nodes and use them to construct $k$ negative edges jointly. Specifically, we reuse these nodes and randomly pair them with nodes in our positive set to generate negative pairs. As a result, we can significantly reduce the number of nodes in a mini-batch and accelerate training. The default method generates $2 \times n + k \times n$ end-point nodes and their neighbor nodes, while the joint negative sampling generates $3 \times n$ end-point nodes.

\textbf{Distributed GNN training}. Finally, to allow scallability to billion node graphs we exploit and extend the distributed GNN training framework~\cite{zheng2021distributed} to accomodate our end-to-end fine-tuning setting. To increase the training efficiency, we apply hierarchical graph partitioning in DGL’s distributed training. When using this method, the target nodes/edges are sampled from the same partition. Therefore, when we sample their neighbors, it’s more likely that different target nodes may sample the same neighbor nodes and thus, reduce the number of nodes in a mini-batch. 

\section{Experimental setting}
\label{sec:exp}

In the experiments we want to evaluate our techniques for improving the transformer representation with graph information to allow better joint fine-tuning with any GNN model. Hence, although \sys can include any GNN model as an encoder here we will only evaluate the GraphSAGE~\cite{hamilton2017representation} for homogenous graphs and the RGCN~\cite{schlichtkrull2018modeling} for heterogenous graphs.

\subsection{Datasets.}

\textbf{Public.} Our unique setting requires graph datasets where the nodes are associated with text. We employ the arxiv, and products datesets from the OGB benchmark~\cite{hu2020open} with $N$=169,343 and $E=$1,166,243 and $N$=2,449,029 $E$=61,859,140 respectively with the standard split ratios from~\cite{hu2020open}. We further augment the data with the original text features for each node; the data are collected in~\cite{chien2021node}. In the arxiv dataset the original title and abstract is used as text feature for the node. On the other hand the product dataset represents Amazon products and the product title was crawled from the web and used as the text feature. The benchmark in~\cite{hu2020open} provides text embeddings of the original text as features for the nodes.  
The task here is to predict the labels on the nodes in a standard semi-supervised setting. The labels are the type the paper and the category of product for arxiv and product respectively. For these datasets we also formulate a link prediction problem with splits 80\% training, 10\% validation and 10\% testing and the tasks are predicting paper citation and product co-purchase links. 
Further we also construct the Yelp dataset augmented with the text using sources provided from~\cite{Yelpdata}. The following node types are included  with corresponding number of nodes business $N_1$=160,585, category $N_2$=1330, city $N_3$=836, review $N_4$=8,635,403, user $N_5$=2,189,457. The following edges are considered (user, friendship, user) $E_1$=17,971,548, (business, in, city) $E_2$=160585, (business, in category, category) $E_3$=708968, (review, on, business) $E_4$=8,635,403, (user, write, review) $E_5$=8,635,403. In this dataset only the review nodes are associated with text. The task here is to predict the stars for a business and is formulated as a node prediction task

\textbf{Private.} Additionally, we consider 
the dataset provided by the recent Amazon KDD22 challenge~\cite{AmazonKDD22}. The graph structure is indeed similar to the one depicted in Fig.~\ref{fig:querygraph}. 
There are $N_1=646,640$ product and $N_2=33,804$ query nodes in the graph and $E=781,744$ edges that represent a match among the query and the product.
Each edge in this dataset is associated with a label which corresponds to whether 
a match between the query and the product is an exact, substitute, complement or irrelevant.
This problem is known as ESCI and is solved as an edge classification task. 
We create a custom split for this task by splitting the set of edges to 60\% for training 10\% for validation and 30\% for testing. 
Finally, we also consider the private query-purchase-product dataset that is used to predict which product will be bought based on a query. Specifically, there are $N_1=130,191,253$ products and $N_2=5,878,377$ queries. Also there are the following behavioral data represented as edges among queries and products: Based on a query a product was added in the basked  $E_1=$15,163,234, a product was clicked $E_2=$24,896,086, a product was returned as candidate $E_3=$140,008,811, was purchased $E_4=$13,138,944. The task here is given a query return the most probable product to be purchased. A natural formulation for this task is under the link prediction setting.

\subsection{Baseline setting}
Next we explain the different parameters that define the various approaches considered in this work.

\noindent\textbf{Encoders}. We consider the following possible semantic encoders in this work. \textsc{BERT} is the pre-trained BERT model from~\cite{wolf2019huggingface}. \textsc{graph-aware BERT} is the pre-trained BERT model~\cite{wolf2019huggingface} that we further fine-tune it for graph structure prediction as in equation~\eqref{eq:rgcnsup}. BERT-PR is a BERT model that is pre-trained using the MLM objective in the proprietary data of the company.\\
\noindent\textbf{Task encoders}. We consider 2 candidate encoders for the experiments presented here. \textsc{MLP} is a single-layer MLP that projects the text embeding to an appropriate dimension for node classification or for link prediction and allows us to circumvent to directly compare with the language model. This is used as a baseline approach to directly use the semantic encoder model in the downstream tasks. \textsc{GraphSAGE} is the model presented in~\cite{graphsage} and is used as our baseline GNN model. \\
\noindent\textbf{Fine-tune}. This parameter defines whether we will back-propagate the loss to the semantic encoder during learning or not. The training is orders of magnitude faster when the loss is not back-propagated.\\
\noindent\textbf{Warm-start}. This parameter defines whether we will warm-start the GNN model by keeping the semantic encoder parameters fixed for some iterations before end-to-end fine-tuning.\\
\noindent\textbf{Model configuration}. We optimize the parameters such that the validation set performance is optimized. We select the number of GNN layers from $1, 2, 3$, GNN hidden dimension from $128, 256, 512$ and learning rate from $10^{-3}, 10^{-4}, 10^{-5}$.

\section{Experiments}

\subsection{Node classification}

Table~\ref{tab:arxiv} collects the results for the public datasets for different training configurations and encoder models in node classification. Notice that the first two rows apply the node prediction loss  directly  on the node representation of the text embeddings after it is appropriately mapped by a single layer MLP. Fine-tuning in this context means that the gradient updates the parameters of the semantic encoder otherwise it is not besides the MLP parameters. The target node for the Yelp dataset does not have any text hence, we can not evaluate the first two settings for that. For this experiment the warm-start did not give significant improvement and hence was not included.

By comparing lines 2 and 3 we observe that fine-tuning the BERT directly for the downstream tasks and disregarding the graph structure  achieves on-par performance as the one of keeping the BERT model fixed and using these representations as input to the GNN model. This suggests that the initial BERT embeddings are indeed not the most appropriate semantic embeddings. 
By comparing lines 3 and 4 we see a performance benefit of fine-tuning the BERT model through the GNN, since the multi-hop information is captured by the GNN.
By comparing lines 3 and 5 we observe the clear advantage of the graph-aware BERT. The graph-aware pre-fine-tuning fuses the transformer with graph information and is the most suitable semantic encoder. 
Finally, line 6 coincides with the proposed \sys framework. We observe that this configuration achieves the best overall performance and includes the proposed stage-wise fine-tuning approach. Hence, fine-tuning the BERT model for link prediction provides good performance in the node classification tasks. This result is very important since it allows to train a BERT model on link prediction and transfer the knowledge on different downstream tasks which may speed up the overall training.

\begin{table*}[t!]
\caption{Node classification results for the public datasets. Results measured in accuracy.}
\vspace{0.1cm}
\centering
\setlength{\tabcolsep}{2.5pt}
\label{tab:arxiv}
\begin{tabular}{llHlccHc}
\toprule
 Semantic encoder& Graph encoder & Warm-start & Fine-tune &
  {arxiv} &
  {products} & {Yelp} & {Yelp}\\ \midrule
\rownumber~~ BERT&MLP & No &No &
  { 62.91} &
  {61.83}  & - & -\\ 
 \rownumber~~ BERT&MLP &No & Yes &
  { 72.98} &
  { 77.64} & - & - \\ 
\rownumber~~  BERT&GNN &No &No &
  {71.39 } &
  79.10 & 61.13 & 65.81 \\ 
\rownumber~~ BERT&GNN &No &Yes &
  {73.42} &
  {81.24} & 66.36 & 73.06 \\ 
\rownumber~~ graph-aware  BERT&GNN &No &No&
  {73.79} &
  {80.53} & 67.51 & 66.88 \\ 
\rownumber~~  graph-aware  BERT&GNN &No &Yes&
  \textbf{74.97} &
  \textbf{82.35} & 60.92 & \textbf{76.47}\\ 
  \bottomrule
  \vspace{-2em}
\end{tabular}
\vspace{-0.2cm}
\end{table*}

\subsection{Link prediction}

Table~\ref{tab:lp} collects  the link prediction performance of the various baselines measured using the MRR score. The first row applies the link prediction supervision in~\eqref{eq:rgcnsup} directly  on the node representation of the text encoding after mapped by a single layer MLP to an embedding and the whole architecture is fine-tuned end-to-end. This model corresponds to the graph-aware pre-fine-tuned model for link prediction and is the same as the used as a semantic encoder in lines 5 and 6. 

By comparing lines 1 versus 2, 3, and 4 we observe that the original BERT model is indeed not appropriate as a semantic encoder used with the GNN. On the other hand, fine-tuning the BERT model for link prediction in row 1 achieves a very good MRR performance.
Lines 5, 6, 7 relative to 2, 3, 4 showcase the advantage of using the graph-aware pre-fine-tuning as an essential step in our \sys framework, where the former leads to a large performance boost. Furthermore, by comparing 5 with 6 and 7 we observe the necessity of warm-starting in certain cases of the GNN encoder to avoid non-desirable local minima. Since the GNN model is initialized at random and the graph-aware BERT is well-trained  optimizing this model without warm-start is challenging.
By comparing lines 3 and 4 we see a performance benefit of fine-tuning the BERT model through the GNN, since the multi-hop information is captured by the GNN.

\textbf{Convergence improvement}. The warm-starting option behinds performance gains in Table 2 it provides significant training speed up. Specifically, for the ogbn-products dataset it takes 168 hours for the option without warm-start (row 6) to reach the maximum performance reported, whereas for the option with warm-start (row 7) it takes only 13 hours to reach the same MRR. Thus warm-start provides a 13x speed up in training speed.
\begin{table*}[t!]
\caption{Link prediction results for the public datasets. The performance is measured in MRR scores. Note that the converged model for row 1  is the graph-aware BERT used in row 5, 6, 7.
}
\vspace{0.1cm}
\centering
\setlength{\tabcolsep}{2.5pt}
\label{tab:lp}
\begin{tabular}{llllcc}
\toprule
 Semantic encoder& Graph encoder & Warm-start & Fine-tune &
  {arxiv} &
  {products} \\ \midrule
\rownumber~~ graph-aware  BERT&MLP & No &Yes &
  { 59.32} &
  {82.29}   \\ 
\rownumber~~  BERT&GNN &No &No &
  {12.43} &
  74.50\\ 
 \rownumber~~  BERT&GNN &No &Yes &
  {10.11} &
  72.13\\ 
  \rownumber~~  BERT&GNN &Yes &Yes &
  {15.23} &
  77.42\\ 
\rownumber~~ graph-aware  BERT&GNN &No &No&
  {58.13} &
  {84.34} \\ 
\rownumber~~  graph-aware  BERT&GNN &No &Yes&
  {55.32} &
  {78.31} \\ 
  \rownumber~~  graph-aware  BERT&GNN &Yes &Yes&
  \textbf{63.21} &
  \textbf{87.23} \\ 
  \bottomrule
  \vspace{-2em}
\end{tabular}
\vspace{-0.2cm}
\end{table*}

\subsection{Public ESCI: edge classification}
The Table~\ref{tab:escipub} contains the edge classification results for the various baselines in ESCI. Note that here we also report the performance for each individual class since we are interested in predicting well also for the rare classes in our application.

By comparing rows 2 and 4 that both fine-tune the BERT-PR model we observe a strong boost of 320 bps in performance when the GNN is used.
This indicates that the GNN can indeed help boosting the performance probably for the rare classes (S-C-I) by a large extend. 
By comparing rows 3 and 4 we observe that it is very important to fine-tune the BERT-PR model during GNN training. 
By comparing rows 3 and 6 we observe that the graph-aware pre-fine-tuning is giving a significant boost when the BERT embeddings are fixed. This benefit diminishes when the BERT model is fine-tuned. 
Finally, the performance in lines 5-8 is quite similar, but interesting the performance in the rare classes is maximized in rows 5 and 6. We plan to dive deeper into these results and analyze the performance for different sample sizes besides the current split.

\begin{table*}[t!]
\caption{Edge classification results for the public ESCI dataset. The performance is measured in F1-score, all classes reported. The converged model in row 2 is the graph-aware BERT used in rows 6,7,8. 
}
\vspace{0.1cm}
\centering
\setlength{\tabcolsep}{2.5pt}
\label{tab:escipub}
\begin{tabular}{llllccccc}
\toprule
 Semantic encoder& Graph encoder & Warm-start & Fine-tune &
  {EvSvCvI} & E & S & C & I \\ \midrule
  \rownumber~~ BERT-PR&MLP & No &No &
  {32.62} & 57.42& 38.12& 14.21 & 20.56 \\
\rownumber~~ graph-aware  BERT&MLP & No &Yes &
  {37.36} & 59.25& 36.13& 21.23& 32.34\\ 
 \rownumber~~  BERT-PR&GNN &No &No &
  {35.12} & 53.34& 38.21& 25.23& 25.42\\
 \rownumber~~  BERT-PR&GNN &No &Yes &
  \textbf{40.56} & \textbf{61.90}& 42.21& {26.30}& {30.21}  \\ 
 \rownumber~~  BERT-PR&GNN &Yes &Yes &
  {40.23} & 60.21& \textbf{45.21}& 30.02& 30.11\\ 
\rownumber~~ graph-aware  BERT&GNN &No &No&
  {37.43} & 55.62&   43.22&  \textbf {32.60}& \textbf {35.24}\\ 
\rownumber~~  graph-aware  BERT&GNN &No &Yes&
  {39.51}  & 60.81& 42.45& 21.56& 28.80\\ 
  \rownumber~~  graph-aware  BERT&GNN &Yes &Yes&
 {40.13} & 60.90& 43.82& 18.20& {31.52}\\ 
  \bottomrule
  \vspace{-2em}
\end{tabular}
\vspace{-0.2cm}
\end{table*}

\subsection{Query-purchase-product dataset}
Table~\ref{tab:semantic} collects the results for predicting if the search query leads to the purchase of a product and is treated as a link prediction task. The evaluation metric is the Macro recall at 100, which is the percentage of true products that exist in the top 100 retrieved products by each method. The results for fine-tuning the BERT model via the GNN model are ongoing.
By comparing rows 1 and 3 that do not fine-tune the BERT-PR model we observe a very large performance boost by the GNN model that considers the graph structure. An even larger performance boost is observed when fine-tuning the BERT model via graph information in row 2. The best performance is observed by using the graph-aware BERT model as a fixed semantic-encoder for the GNN model. Future steps will focus on end-to-end fine-tuning. 


\begin{table*}[t!]
\caption{Macro recall at 100 for the query-purchase-product dataset. The converged model for row 2 is the graph-aware BERT in row 4.
}
\vspace{0.1cm}
\centering
\setlength{\tabcolsep}{2.5pt}
\label{tab:semantic}
\begin{tabular}{llHlc}
\toprule
 Semantic encoder& Graph encoder & Warm-start & Fine-tune &
  {Macro@100}  \\ \midrule
  \rownumber~~ BERT-PR&MLP & No &No &
  {34.12} \\
\rownumber~~ graph-aware  BERT&MLP & No &Yes &
  {79.06} \\ 
 \rownumber~~  BERT-PR&GNN &No &No &
  {77.26}\\
\rownumber~~ graph-aware  BERT&GNN &No &No&
  \textbf{86.53} \\ 
  \bottomrule
  \vspace{-2em}
\end{tabular}
\vspace{-0.2cm}
\end{table*}

\section{Conclusion}
In this paper we develop a framework termed \sys that achieves high-quality representations for graph data with rich textual features. Our framework employs stage-wise fine-tuning steps that allow for the BERT model to gradually adapt to the graph domain data. We prove with experiments in four public datasets and one query-purchase-product dataset the power of the \sys framework.


\bibliographystyle{plain}
\bibliography{my_bibliography.bib}
\appendix
\end{document}

%% file: figures/embedding_analysis/basic_graph.tex
\tikzset{every picture/.style={line width=0.75pt}} 

\begin{tikzpicture}[x=0.75pt,y=0.75pt,yscale=-1,xscale=1]

\draw  [fill=p1  ,fill opacity=1 ] (323,86) .. controls (323,80.48) and (327.48,76) .. (333,76) .. controls (338.52,76) and (343,80.48) .. (343,86) .. controls (343,91.52) and (338.52,96) .. (333,96) .. controls (327.48,96) and (323,91.52) .. (323,86) -- cycle ;
\draw  [fill=p3  ,fill opacity=1 ] (329,210) .. controls (329,204.48) and (333.48,200) .. (339,200) .. controls (344.52,200) and (349,204.48) .. (349,210) .. controls (349,215.52) and (344.52,220) .. (339,220) .. controls (333.48,220) and (329,215.52) .. (329,210) -- cycle ;
\draw  [fill=p1  ,fill opacity=1 ] (334,318) .. controls (334,312.48) and (338.48,308) .. (344,308) .. controls (349.52,308) and (354,312.48) .. (354,318) .. controls (354,323.52) and (349.52,328) .. (344,328) .. controls (338.48,328) and (334,323.52) .. (334,318) -- cycle ;
\draw    (164,202.5) .. controls (190.73,133.2) and (281.17,116.34) .. (321.78,86.9) ;
\draw [shift={(323,86)}, rotate = 143.13] [color={rgb, 255:red, 0; green, 0; blue, 0 }  ][line width=0.75]    (10.93,-3.29) .. controls (6.95,-1.4) and (3.31,-0.3) .. (0,0) .. controls (3.31,0.3) and (6.95,1.4) .. (10.93,3.29)   ;
\draw    (164,208) .. controls (203.6,178.3) and (292.21,344.62) .. (332.79,318.84) ;
\draw [shift={(334,318)}, rotate = 143.13] [color={rgb, 255:red, 0; green, 0; blue, 0 }  ][line width=0.75]    (10.93,-3.29) .. controls (6.95,-1.4) and (3.31,-0.3) .. (0,0) .. controls (3.31,0.3) and (6.95,1.4) .. (10.93,3.29)   ;
\draw    (167,84) .. controls (193.73,14.7) and (287.11,235.51) .. (327.78,210.82) ;
\draw [shift={(329,210)}, rotate = 143.13] [color={rgb, 255:red, 0; green, 0; blue, 0 }  ][line width=0.75]    (10.93,-3.29) .. controls (6.95,-1.4) and (3.31,-0.3) .. (0,0) .. controls (3.31,0.3) and (6.95,1.4) .. (10.93,3.29)   ;
\draw    (401,248) .. controls (440.6,218.3) and (460.6,276.81) .. (499.81,248.87) ;
\draw [shift={(501,248)}, rotate = 143.13] [color={rgb, 255:red, 0; green, 0; blue, 0 }  ][line width=0.75]    (10.93,-3.29) .. controls (6.95,-1.4) and (3.31,-0.3) .. (0,0) .. controls (3.31,0.3) and (6.95,1.4) .. (10.93,3.29)   ;
\draw  [fill=p1  ,fill opacity=1 ] (433,193) .. controls (433,187.48) and (437.48,183) .. (443,183) .. controls (448.52,183) and (453,187.48) .. (453,193) .. controls (453,198.52) and (448.52,203) .. (443,203) .. controls (437.48,203) and (433,198.52) .. (433,193) -- cycle ;
\draw  [fill=q2  ,fill opacity=1 ] (158.5,80.5) -- (169,99) -- (148,99) -- cycle ;
\draw  [fill=q1  ,fill opacity=1 ] (158.5,197.5) -- (169,216) -- (148,216) -- cycle ;
\draw  [fill=q1,fill opacity=1 ] (441.5,133.5) -- (452,152) -- (431,152) -- cycle ;

\draw (416,267) node [anchor=north west][inner sep=0.75pt]   [align=left] {Purchased};
\draw (116,232) node [anchor=north west][inner sep=0.75pt]   [align=left] {Running shoes};
\draw (118,104) node [anchor=north west][inner sep=0.75pt]   [align=left] {Hiking shoes};
\draw (276,112) node [anchor=north west][inner sep=0.75pt]   [align=left] {Nike ZoomX Vaporfly};
\draw (302,344) node [anchor=north west][inner sep=0.75pt]   [align=left] {Addidas swift};
\draw (424,212) node [anchor=north west][inner sep=0.75pt]   [align=left] {Product};
\draw (421,157) node [anchor=north west][inner sep=0.75pt]   [align=left] {Query};
\draw (282,230) node [anchor=north west][inner sep=0.75pt]   [align=left] {Timberland boots};

\end{tikzpicture}

%% file: figures/embedding_analysis/tembed.tex
\tikzset{every picture/.style={line width=0.75pt}} 

\begin{tikzpicture}[x=0.75pt,y=0.75pt,yscale=-1,xscale=1]

\draw  [fill=p1  ,fill opacity=1 ] (315,77) .. controls (315,71.48) and (319.48,67) .. (325,67) .. controls (330.52,67) and (335,71.48) .. (335,77) .. controls (335,82.52) and (330.52,87) .. (325,87) .. controls (319.48,87) and (315,82.52) .. (315,77) -- cycle ;
\draw  [fill=p3  ,fill opacity=1 ] (220,219) .. controls (220,213.48) and (224.48,209) .. (230,209) .. controls (235.52,209) and (240,213.48) .. (240,219) .. controls (240,224.52) and (235.52,229) .. (230,229) .. controls (224.48,229) and (220,224.52) .. (220,219) -- cycle ;
\draw  [fill=p1  ,fill opacity=1 ] (384,321) .. controls (384,315.48) and (388.48,311) .. (394,311) .. controls (399.52,311) and (404,315.48) .. (404,321) .. controls (404,326.52) and (399.52,331) .. (394,331) .. controls (388.48,331) and (384,326.52) .. (384,321) -- cycle ;
\draw  (50,378.15) -- (486,378.15)(93.6,114) -- (93.6,407.5) (479,373.15) -- (486,378.15) -- (479,383.15) (88.6,121) -- (93.6,114) -- (98.6,121)  ;
\draw  [fill=q2  ,fill opacity=1 ] (168.5,174.5) -- (179,193) -- (158,193) -- cycle ;
\draw  [fill=q1  ,fill opacity=1 ] (161.5,232.5) -- (172,251) -- (151,251) -- cycle ;

\draw (112,255) node [anchor=north west][inner sep=0.75pt]   [align=left] {Running shoes};
\draw (124,197) node [anchor=north west][inner sep=0.75pt]   [align=left] {Hiking shoes};
\draw (269,103) node [anchor=north west][inner sep=0.75pt]   [align=left] {Nike ZoomX Vaporfly};
\draw (352,347) node [anchor=north west][inner sep=0.75pt]   [align=left] {Addidas swift};
\draw (197,239) node [anchor=north west][inner sep=0.75pt]   [align=left] {Timberland boots};

\end{tikzpicture}

%% file: figures/embedding_analysis/gtembed.tex
\tikzset{every picture/.style={line width=0.75pt}} 

\begin{tikzpicture}[x=0.75pt,y=0.75pt,yscale=-1,xscale=1]

\draw  [fill=p1  ,fill opacity=1 ] (297,168) .. controls (297,162.48) and (301.48,158) .. (307,158) .. controls (312.52,158) and (317,162.48) .. (317,168) .. controls (317,173.52) and (312.52,178) .. (307,178) .. controls (301.48,178) and (297,173.52) .. (297,168) -- cycle ;
\draw  [fill=p3  ,fill opacity=1 ] (160,295) .. controls (160,289.48) and (164.48,285) .. (170,285) .. controls (175.52,285) and (180,289.48) .. (180,295) .. controls (180,300.52) and (175.52,305) .. (170,305) .. controls (164.48,305) and (160,300.52) .. (160,295) -- cycle ;
\draw  [fill=p1  ,fill opacity=1 ] (355,343) .. controls (355,337.48) and (359.48,333) .. (365,333) .. controls (370.52,333) and (375,337.48) .. (375,343) .. controls (375,348.52) and (370.52,353) .. (365,353) .. controls (359.48,353) and (355,348.52) .. (355,343) -- cycle ;
\draw  (62,400.25) -- (479,400.25)(103.7,128) -- (103.7,430.5) (472,395.25) -- (479,400.25) -- (472,405.25) (98.7,135) -- (103.7,128) -- (108.7,135)  ;
\draw  [fill=q2  ,fill opacity=1 ] (252.5,277.5) -- (263,296) -- (242,296) -- cycle ;
\draw  [fill=q1  ,fill opacity=1 ] (320.5,243.5) -- (331,262) -- (310,262) -- cycle ;

\draw (295,266) node [anchor=north west][inner sep=0.75pt]   [align=left] {Running shoes};
\draw (228,301) node [anchor=north west][inner sep=0.75pt]   [align=left] {Hiking shoes};
\draw (251,192) node [anchor=north west][inner sep=0.75pt]   [align=left] {Nike ZoomX Vaporfly};
\draw (359,359) node [anchor=north west][inner sep=0.75pt]   [align=left] {Addidas swift};
\draw (115,312) node [anchor=north west][inner sep=0.75pt]   [align=left] {Timberland boots};

\end{tikzpicture}

%% file: figures/embedding_analysis/gnnembed.tex
\tikzset{every picture/.style={line width=0.75pt}} 

\begin{tikzpicture}[x=0.75pt,y=0.75pt,yscale=-1,xscale=1]

\draw  [fill=p1  ,fill opacity=1 ] (310,211) .. controls (310,205.48) and (314.48,201) .. (320,201) .. controls (325.52,201) and (330,205.48) .. (330,211) .. controls (330,216.52) and (325.52,221) .. (320,221) .. controls (314.48,221) and (310,216.52) .. (310,211) -- cycle ;
\draw  [fill=p3  ,fill opacity=1 ] (124,317) .. controls (124,311.48) and (128.48,307) .. (134,307) .. controls (139.52,307) and (144,311.48) .. (144,317) .. controls (144,322.52) and (139.52,327) .. (134,327) .. controls (128.48,327) and (124,322.52) .. (124,317) -- cycle ;
\draw  [fill=p1  ,fill opacity=1 ] (347,269) .. controls (347,263.48) and (351.48,259) .. (357,259) .. controls (362.52,259) and (367,263.48) .. (367,269) .. controls (367,274.52) and (362.52,279) .. (357,279) .. controls (351.48,279) and (347,274.52) .. (347,269) -- cycle ;
\draw  (12,415.95) -- (433,415.95)(54.1,159) -- (54.1,444.5) (426,410.95) -- (433,415.95) -- (426,420.95) (49.1,166) -- (54.1,159) -- (59.1,166)  ;
\draw  [fill=q2  ,fill opacity=1 ] (189.5,296.5) -- (200,315) -- (179,315) -- cycle ;
\draw  [fill=q1  ,fill opacity=1 ] (259.5,274.5) -- (270,293) -- (249,293) -- cycle ;

\draw (209,304) node [anchor=north west][inner sep=0.75pt]   [align=left] {Running shoes};
\draw (162,325) node [anchor=north west][inner sep=0.75pt]   [align=left] {Hiking shoes};
\draw (265,236) node [anchor=north west][inner sep=0.75pt]   [align=left] {Nike ZoomX Vaporfly};
\draw (315,295) node [anchor=north west][inner sep=0.75pt]   [align=left] {Addidas swift};
\draw (85,342) node [anchor=north west][inner sep=0.75pt]   [align=left] {Timberland boots};

\end{tikzpicture}

%% file: figures/models/gtrans.tex
  
\tikzset {_g3f0u153l/.code = {\pgfsetadditionalshadetransform{ \pgftransformshift{\pgfpoint{0 bp } { 0 bp }  }  \pgftransformrotate{0 }  \pgftransformscale{2 }  }}}
\pgfdeclarehorizontalshading{_zy9cub4rb}{150bp}{rgb(0bp)=(1,1,1);
rgb(37.5bp)=(1,1,1);
rgb(37.5bp)=(0.82,0.92,0.98);
rgb(100bp)=(0.82,0.92,0.98)}

  
\tikzset {_il8l5ay61/.code = {\pgfsetadditionalshadetransform{ \pgftransformshift{\pgfpoint{0 bp } { 0 bp }  }  \pgftransformrotate{0 }  \pgftransformscale{2 }  }}}
\pgfdeclarehorizontalshading{_acqbtocqn}{150bp}{rgb(0bp)=(1,1,1);
rgb(37.5bp)=(1,1,1);
rgb(37.5bp)=(0.82,0.92,0.98);
rgb(100bp)=(0.82,0.92,0.98)}

  
\tikzset {_blvg7mxy6/.code = {\pgfsetadditionalshadetransform{ \pgftransformshift{\pgfpoint{0 bp } { 0 bp }  }  \pgftransformrotate{0 }  \pgftransformscale{2 }  }}}
\pgfdeclarehorizontalshading{_6ngmyd4fx}{150bp}{rgb(0bp)=(1,0.8,0.69);
rgb(37.5bp)=(1,0.8,0.69);
rgb(50bp)=(0.95,0.45,0.2);
rgb(60.69614955357143bp)=(0.92,0.33,0.03);
rgb(62.5bp)=(0.98,0.58,0.37);
rgb(100bp)=(0.98,0.58,0.37)}

  
\tikzset {_jkj08w2yj/.code = {\pgfsetadditionalshadetransform{ \pgftransformshift{\pgfpoint{0 bp } { 0 bp }  }  \pgftransformrotate{0 }  \pgftransformscale{2 }  }}}
\pgfdeclarehorizontalshading{_wffsqh5yn}{150bp}{rgb(0bp)=(1,0.8,0.69);
rgb(37.5bp)=(1,0.8,0.69);
rgb(50bp)=(0.95,0.45,0.2);
rgb(60.69614955357143bp)=(0.92,0.33,0.03);
rgb(62.5bp)=(0.98,0.58,0.37);
rgb(100bp)=(0.98,0.58,0.37)}

  
\tikzset {_sx107gsxt/.code = {\pgfsetadditionalshadetransform{ \pgftransformshift{\pgfpoint{0 bp } { 0 bp }  }  \pgftransformrotate{0 }  \pgftransformscale{2 }  }}}
\pgfdeclarehorizontalshading{_n7ppu8fe5}{150bp}{rgb(0bp)=(1,0.8,0.69);
rgb(37.5bp)=(1,0.8,0.69);
rgb(50bp)=(0.95,0.45,0.2);
rgb(60.69614955357143bp)=(0.92,0.33,0.03);
rgb(62.5bp)=(0.98,0.58,0.37);
rgb(100bp)=(0.98,0.58,0.37)}
\tikzset{every picture/.style={line width=0.75pt}} 

\begin{tikzpicture}[x=0.75pt,y=0.75pt,yscale=-1,xscale=1]

\draw  [fill=p1  ,fill opacity=1 ] (575,71) .. controls (575,65.48) and (579.48,61) .. (585,61) .. controls (590.52,61) and (595,65.48) .. (595,71) .. controls (595,76.52) and (590.52,81) .. (585,81) .. controls (579.48,81) and (575,76.52) .. (575,71) -- cycle ;
\draw    (512,117.5) .. controls (538.73,48.2) and (535.08,99.94) .. (573.81,71.88) ;
\draw [shift={(575,71)}, rotate = 143.13] [color={rgb, 255:red, 0; green, 0; blue, 0 }  ][line width=0.75]    (10.93,-3.29) .. controls (6.95,-1.4) and (3.31,-0.3) .. (0,0) .. controls (3.31,0.3) and (6.95,1.4) .. (10.93,3.29)   ;
\path  [shading=_zy9cub4rb,_g3f0u153l] (144,60.8) .. controls (144,54.56) and (149.06,49.5) .. (155.3,49.5) -- (272.7,49.5) .. controls (278.94,49.5) and (284,54.56) .. (284,60.8) -- (284,94.7) .. controls (284,100.94) and (278.94,106) .. (272.7,106) -- (155.3,106) .. controls (149.06,106) and (144,100.94) .. (144,94.7) -- cycle ; 
 \draw   (144,60.8) .. controls (144,54.56) and (149.06,49.5) .. (155.3,49.5) -- (272.7,49.5) .. controls (278.94,49.5) and (284,54.56) .. (284,60.8) -- (284,94.7) .. controls (284,100.94) and (278.94,106) .. (272.7,106) -- (155.3,106) .. controls (149.06,106) and (144,100.94) .. (144,94.7) -- cycle ; 

\path  [shading=_acqbtocqn,_il8l5ay61] (330,60.8) .. controls (330,54.56) and (335.06,49.5) .. (341.3,49.5) -- (458.7,49.5) .. controls (464.94,49.5) and (470,54.56) .. (470,60.8) -- (470,94.7) .. controls (470,100.94) and (464.94,106) .. (458.7,106) -- (341.3,106) .. controls (335.06,106) and (330,100.94) .. (330,94.7) -- cycle ; 
 \draw   (330,60.8) .. controls (330,54.56) and (335.06,49.5) .. (341.3,49.5) -- (458.7,49.5) .. controls (464.94,49.5) and (470,54.56) .. (470,60.8) -- (470,94.7) .. controls (470,100.94) and (464.94,106) .. (458.7,106) -- (341.3,106) .. controls (335.06,106) and (330,100.94) .. (330,94.7) -- cycle ; 

\draw  [fill=q1  ,fill opacity=1 ] (505.5,106.5) -- (516,125) -- (495,125) -- cycle ;
\path  [shading=_6ngmyd4fx,_blvg7mxy6] (110,69.13) -- (117.8,69.13) -- (117.8,64.5) -- (123,73.75) -- (117.8,83) -- (117.8,78.38) -- (110,78.38) -- cycle ; 
 \draw   (110,69.13) -- (117.8,69.13) -- (117.8,64.5) -- (123,73.75) -- (117.8,83) -- (117.8,78.38) -- (110,78.38) -- cycle ; 

\path  [shading=_wffsqh5yn,_jkj08w2yj] (302,71.13) -- (309.8,71.13) -- (309.8,66.5) -- (315,75.75) -- (309.8,85) -- (309.8,80.38) -- (302,80.38) -- cycle ; 
 \draw   (302,71.13) -- (309.8,71.13) -- (309.8,66.5) -- (315,75.75) -- (309.8,85) -- (309.8,80.38) -- (302,80.38) -- cycle ; 

\path  [shading=_n7ppu8fe5,_sx107gsxt] (489,73.13) -- (496.8,73.13) -- (496.8,68.5) -- (502,77.75) -- (496.8,87) -- (496.8,82.38) -- (489,82.38) -- cycle ; 
 \draw   (489,73.13) -- (496.8,73.13) -- (496.8,68.5) -- (502,77.75) -- (496.8,87) -- (496.8,82.38) -- (489,82.38) -- cycle ; 

\draw (481,132) node [anchor=north west][inner sep=0.75pt]  [font=\footnotesize] [align=left] {Running shoes};
\draw (561,44) node [anchor=north west][inner sep=0.75pt]  [font=\footnotesize] [align=left] {Nike ZoomX Vaporfly};
\draw (175,60) node [anchor=north west][inner sep=0.75pt]   [align=left] {Graph-aware\\transformer};
\draw (335,61) node [anchor=north west][inner sep=0.75pt]   [align=left] {\begin{minipage}[lt]{91.17pt}\setlength\topsep{0pt}
\begin{center}
Structure prediction\\decoder
\end{center}

\end{minipage}};
\draw (4,58) node [anchor=north west][inner sep=0.75pt]  [font=\footnotesize] [align=left] {Running shoes};
\draw (6,90) node [anchor=north west][inner sep=0.75pt]  [font=\footnotesize] [align=left] {Nike ZoomX Vaporfly};
\draw (529,70) node [anchor=north west][inner sep=0.75pt]  [color=q1  ,opacity=1 ] [align=left] {{\Large ?}};

\end{tikzpicture}

%% file: figures/models/lmgnn.tex
  
\tikzset {_961w7o3r2/.code = {\pgfsetadditionalshadetransform{ \pgftransformshift{\pgfpoint{0 bp } { 0 bp }  }  \pgftransformrotate{0 }  \pgftransformscale{2 }  }}}
\pgfdeclarehorizontalshading{_749yt98j6}{150bp}{rgb(0bp)=(1,1,1);
rgb(37.5bp)=(1,1,1);
rgb(37.5bp)=(0.82,0.92,0.98);
rgb(100bp)=(0.82,0.92,0.98)}

  
\tikzset {_quw5uiiap/.code = {\pgfsetadditionalshadetransform{ \pgftransformshift{\pgfpoint{0 bp } { 0 bp }  }  \pgftransformrotate{0 }  \pgftransformscale{2 }  }}}
\pgfdeclarehorizontalshading{_qr3wt5wpf}{150bp}{rgb(0bp)=(1,1,1);
rgb(37.5bp)=(1,1,1);
rgb(37.5bp)=(0.82,0.92,0.98);
rgb(100bp)=(0.82,0.92,0.98)}

  
\tikzset {_0mly69xkx/.code = {\pgfsetadditionalshadetransform{ \pgftransformshift{\pgfpoint{0 bp } { 0 bp }  }  \pgftransformrotate{0 }  \pgftransformscale{2 }  }}}
\pgfdeclarehorizontalshading{_h8itfd31b}{150bp}{rgb(0bp)=(1,0.8,0.69);
rgb(37.5bp)=(1,0.8,0.69);
rgb(50bp)=(0.95,0.45,0.2);
rgb(60.69614955357143bp)=(0.92,0.33,0.03);
rgb(62.5bp)=(0.98,0.58,0.37);
rgb(100bp)=(0.98,0.58,0.37)}

  
\tikzset {_5ju10megz/.code = {\pgfsetadditionalshadetransform{ \pgftransformshift{\pgfpoint{0 bp } { 0 bp }  }  \pgftransformrotate{0 }  \pgftransformscale{2 }  }}}
\pgfdeclarehorizontalshading{_6c1anlnxd}{150bp}{rgb(0bp)=(1,0.8,0.69);
rgb(37.5bp)=(1,0.8,0.69);
rgb(50bp)=(0.95,0.45,0.2);
rgb(60.69614955357143bp)=(0.92,0.33,0.03);
rgb(62.5bp)=(0.98,0.58,0.37);
rgb(100bp)=(0.98,0.58,0.37)}

  
\tikzset {_vty58hx9q/.code = {\pgfsetadditionalshadetransform{ \pgftransformshift{\pgfpoint{0 bp } { 0 bp }  }  \pgftransformrotate{0 }  \pgftransformscale{2 }  }}}
\pgfdeclarehorizontalshading{_c09bfx078}{150bp}{rgb(0bp)=(1,0.8,0.69);
rgb(37.5bp)=(1,0.8,0.69);
rgb(50bp)=(0.95,0.45,0.2);
rgb(60.69614955357143bp)=(0.92,0.33,0.03);
rgb(62.5bp)=(0.98,0.58,0.37);
rgb(100bp)=(0.98,0.58,0.37)}

  
\tikzset {_udxh9wy2a/.code = {\pgfsetadditionalshadetransform{ \pgftransformshift{\pgfpoint{0 bp } { 0 bp }  }  \pgftransformrotate{0 }  \pgftransformscale{2 }  }}}
\pgfdeclarehorizontalshading{_jlho08jne}{150bp}{rgb(0bp)=(1,0.8,0.69);
rgb(37.5bp)=(1,0.8,0.69);
rgb(50bp)=(0.95,0.45,0.2);
rgb(60.69614955357143bp)=(0.92,0.33,0.03);
rgb(62.5bp)=(0.98,0.58,0.37);
rgb(100bp)=(0.98,0.58,0.37)}

  
\tikzset {_r31x0oefc/.code = {\pgfsetadditionalshadetransform{ \pgftransformshift{\pgfpoint{0 bp } { 0 bp }  }  \pgftransformrotate{0 }  \pgftransformscale{2 }  }}}
\pgfdeclarehorizontalshading{_v3g051mqa}{150bp}{rgb(0bp)=(1,1,1);
rgb(37.5bp)=(1,1,1);
rgb(37.5bp)=(0.82,0.92,0.98);
rgb(100bp)=(0.82,0.92,0.98)}

  
\tikzset {_6gk7ddubi/.code = {\pgfsetadditionalshadetransform{ \pgftransformshift{\pgfpoint{0 bp } { 0 bp }  }  \pgftransformrotate{0 }  \pgftransformscale{2 }  }}}
\pgfdeclarehorizontalshading{_8f4upc93w}{150bp}{rgb(0bp)=(1,0.8,0.69);
rgb(37.5bp)=(1,0.8,0.69);
rgb(37.5bp)=(0.92,0.33,0.03);
rgb(50bp)=(0.95,0.45,0.2);
rgb(62.5bp)=(0.98,0.58,0.37);
rgb(100bp)=(0.98,0.58,0.37)}
\tikzset{every picture/.style={line width=0.75pt}} 

\begin{tikzpicture}[x=0.75pt,y=0.75pt,yscale=-1,xscale=1]

\draw  [fill=p1  ,fill opacity=1 ] (735,74) .. controls (735,68.48) and (739.48,64) .. (745,64) .. controls (750.52,64) and (755,68.48) .. (755,74) .. controls (755,79.52) and (750.52,84) .. (745,84) .. controls (739.48,84) and (735,79.52) .. (735,74) -- cycle ;
\draw    (672,120.5) .. controls (698.73,51.2) and (695.08,102.94) .. (733.81,74.88) ;
\draw [shift={(735,74)}, rotate = 143.13] [color={rgb, 255:red, 0; green, 0; blue, 0 }  ][line width=0.75]    (10.93,-3.29) .. controls (6.95,-1.4) and (3.31,-0.3) .. (0,0) .. controls (3.31,0.3) and (6.95,1.4) .. (10.93,3.29)   ;
\path  [shading=_749yt98j6,_961w7o3r2] (151,60.8) .. controls (151,54.56) and (156.06,49.5) .. (162.3,49.5) -- (279.7,49.5) .. controls (285.94,49.5) and (291,54.56) .. (291,60.8) -- (291,94.7) .. controls (291,100.94) and (285.94,106) .. (279.7,106) -- (162.3,106) .. controls (156.06,106) and (151,100.94) .. (151,94.7) -- cycle ; 
 \draw   (151,60.8) .. controls (151,54.56) and (156.06,49.5) .. (162.3,49.5) -- (279.7,49.5) .. controls (285.94,49.5) and (291,54.56) .. (291,60.8) -- (291,94.7) .. controls (291,100.94) and (285.94,106) .. (279.7,106) -- (162.3,106) .. controls (156.06,106) and (151,100.94) .. (151,94.7) -- cycle ; 

\path  [shading=_qr3wt5wpf,_quw5uiiap] (490,63.8) .. controls (490,57.56) and (495.06,52.5) .. (501.3,52.5) -- (618.7,52.5) .. controls (624.94,52.5) and (630,57.56) .. (630,63.8) -- (630,97.7) .. controls (630,103.94) and (624.94,109) .. (618.7,109) -- (501.3,109) .. controls (495.06,109) and (490,103.94) .. (490,97.7) -- cycle ; 
 \draw   (490,63.8) .. controls (490,57.56) and (495.06,52.5) .. (501.3,52.5) -- (618.7,52.5) .. controls (624.94,52.5) and (630,57.56) .. (630,63.8) -- (630,97.7) .. controls (630,103.94) and (624.94,109) .. (618.7,109) -- (501.3,109) .. controls (495.06,109) and (490,103.94) .. (490,97.7) -- cycle ; 

\draw  [fill=q1  ,fill opacity=1 ] (665.5,109.5) -- (676,128) -- (655,128) -- cycle ;
\path  [shading=_h8itfd31b,_0mly69xkx] (117,69.13) -- (124.8,69.13) -- (124.8,64.5) -- (130,73.75) -- (124.8,83) -- (124.8,78.38) -- (117,78.38) -- cycle ; 
 \draw   (117,69.13) -- (124.8,69.13) -- (124.8,64.5) -- (130,73.75) -- (124.8,83) -- (124.8,78.38) -- (117,78.38) -- cycle ; 

\path  [shading=_6c1anlnxd,_5ju10megz] (302,70.13) -- (309.8,70.13) -- (309.8,65.5) -- (315,74.75) -- (309.8,84) -- (309.8,79.38) -- (302,79.38) -- cycle ; 
 \draw   (302,70.13) -- (309.8,70.13) -- (309.8,65.5) -- (315,74.75) -- (309.8,84) -- (309.8,79.38) -- (302,79.38) -- cycle ; 

\path  [shading=_c09bfx078,_vty58hx9q] (649,76.13) -- (656.8,76.13) -- (656.8,71.5) -- (662,80.75) -- (656.8,90) -- (656.8,85.38) -- (649,85.38) -- cycle ; 
 \draw   (649,76.13) -- (656.8,76.13) -- (656.8,71.5) -- (662,80.75) -- (656.8,90) -- (656.8,85.38) -- (649,85.38) -- cycle ; 

\path  [shading=_jlho08jne,_udxh9wy2a] (466,71.13) -- (473.8,71.13) -- (473.8,66.5) -- (479,75.75) -- (473.8,85) -- (473.8,80.38) -- (466,80.38) -- cycle ; 
 \draw   (466,71.13) -- (473.8,71.13) -- (473.8,66.5) -- (479,75.75) -- (473.8,85) -- (473.8,80.38) -- (466,80.38) -- cycle ; 

\path  [shading=_v3g051mqa,_r31x0oefc] (319,59.8) .. controls (319,53.56) and (324.06,48.5) .. (330.3,48.5) -- (447.7,48.5) .. controls (453.94,48.5) and (459,53.56) .. (459,59.8) -- (459,93.7) .. controls (459,99.94) and (453.94,105) .. (447.7,105) -- (330.3,105) .. controls (324.06,105) and (319,99.94) .. (319,93.7) -- cycle ; 
 \draw   (319,59.8) .. controls (319,53.56) and (324.06,48.5) .. (330.3,48.5) -- (447.7,48.5) .. controls (453.94,48.5) and (459,53.56) .. (459,59.8) -- (459,93.7) .. controls (459,99.94) and (453.94,105) .. (447.7,105) -- (330.3,105) .. controls (324.06,105) and (319,99.94) .. (319,93.7) -- cycle ; 

\draw  [fill=p1  ,fill opacity=1 ] (418.36,142.41) .. controls (418.36,140.67) and (420.18,139.25) .. (422.44,139.25) .. controls (424.69,139.25) and (426.51,140.67) .. (426.51,142.41) .. controls (426.51,144.16) and (424.69,145.58) .. (422.44,145.58) .. controls (420.18,145.58) and (418.36,144.16) .. (418.36,142.41) -- cycle ;
\draw  [fill=p3  ,fill opacity=1 ] (420.81,181.66) .. controls (420.81,179.91) and (422.63,178.49) .. (424.88,178.49) .. controls (427.14,178.49) and (428.96,179.91) .. (428.96,181.66) .. controls (428.96,183.4) and (427.14,184.82) .. (424.88,184.82) .. controls (422.63,184.82) and (420.81,183.4) .. (420.81,181.66) -- cycle ;
\draw  [fill=p1  ,fill opacity=1 ] (422.84,215.84) .. controls (422.84,214.09) and (424.67,212.67) .. (426.92,212.67) .. controls (429.17,212.67) and (431,214.09) .. (431,215.84) .. controls (431,217.58) and (429.17,219) .. (426.92,219) .. controls (424.67,219) and (422.84,217.58) .. (422.84,215.84) -- cycle ;
\draw    (353.52,179.28) .. controls (364.2,157.79) and (399.82,152.24) .. (416.84,143.26) ;
\draw [shift={(418.36,142.41)}, rotate = 149.8] [color={rgb, 255:red, 0; green, 0; blue, 0 }  ][line width=0.75]    (10.93,-3.29) .. controls (6.95,-1.4) and (3.31,-0.3) .. (0,0) .. controls (3.31,0.3) and (6.95,1.4) .. (10.93,3.29)   ;
\draw    (353.52,181.02) .. controls (369.26,171.86) and (403.99,221.64) .. (421.06,216.6) ;
\draw [shift={(422.84,215.84)}, rotate = 149.8] [color={rgb, 255:red, 0; green, 0; blue, 0 }  ][line width=0.75]    (10.93,-3.29) .. controls (6.95,-1.4) and (3.31,-0.3) .. (0,0) .. controls (3.31,0.3) and (6.95,1.4) .. (10.93,3.29)   ;
\draw    (354.75,141.78) .. controls (365.37,120.4) and (401.82,186.26) .. (419.01,182.36) ;
\draw [shift={(420.81,181.66)}, rotate = 149.8] [color={rgb, 255:red, 0; green, 0; blue, 0 }  ][line width=0.75]    (10.93,-3.29) .. controls (6.95,-1.4) and (3.31,-0.3) .. (0,0) .. controls (3.31,0.3) and (6.95,1.4) .. (10.93,3.29)   ;
\draw  [fill=q2  ,fill opacity=1 ] (351.28,140.67) -- (355.56,146.53) -- (347,146.53) -- cycle ;
\draw  [fill=q1  ,fill opacity=1 ] (351.28,177.7) -- (355.56,183.55) -- (347,183.55) -- cycle ;
\path  [shading=_8f4upc93w,_6gk7ddubi] (382,118.1) -- (389,112.5) -- (396,118.1) -- (392.5,118.1) -- (392.5,126.5) -- (385.5,126.5) -- (385.5,118.1) -- cycle ; 
 \draw   (382,118.1) -- (389,112.5) -- (396,118.1) -- (392.5,118.1) -- (392.5,126.5) -- (385.5,126.5) -- (385.5,118.1) -- cycle ; 

\draw   (137,13.5) -- (486,13.5) -- (486,138.5) -- (137,138.5) -- cycle ;

\draw (628,148) node [anchor=north west][inner sep=0.75pt]  [font=\footnotesize] [align=left] {Running shoes};
\draw (686,39) node [anchor=north west][inner sep=0.75pt]  [font=\footnotesize] [align=left] {Nike ZoomX Vaporfly};
\draw (182,60) node [anchor=north west][inner sep=0.75pt]   [align=left] {Graph-aware\\transformer};
\draw (495,64) node [anchor=north west][inner sep=0.75pt]   [align=left] {\begin{minipage}[lt]{91.17pt}\setlength\topsep{0pt}
\begin{center}
Structure prediction\\decoder
\end{center}

\end{minipage}};
\draw (11,58) node [anchor=north west][inner sep=0.75pt]  [font=\footnotesize] [align=left] {Running shoes};
\draw (13,90) node [anchor=north west][inner sep=0.75pt]  [font=\footnotesize] [align=left] {Nike ZoomX Vaporfly};
\draw (689,73) node [anchor=north west][inner sep=0.75pt]  [color=q1  ,opacity=1 ] [align=left] {{\Large ?}};
\draw (345,66) node [anchor=north west][inner sep=0.75pt]   [align=left] {GNN encoder};
\draw (418,21) node [anchor=north west][inner sep=0.75pt]   [align=left] {LM-GNN};

\end{tikzpicture}